\documentclass[conference]{IEEEtran}
\IEEEoverridecommandlockouts
\usepackage[colorlinks,
linkcolor=red,
anchorcolor=blue,
citecolor=green]{hyperref}

\usepackage{cite}
\usepackage{amsmath}
\usepackage{algorithmic}
\usepackage{graphicx}
\usepackage{textcomp}
\usepackage{xcolor}
\usepackage{booktabs} 
\usepackage{array}  
\usepackage{multirow}
\usepackage{amssymb}
\usepackage{amsfonts}
\usepackage{bm}
\def\BibTeX{{\rm B\kern-.05em{\sc i\kern-.025em b}\kern-.08em
    T\kern-.1667em\lower.7ex\hbox{E}\kern-.125emX}}

\makeatletter
\newcommand{\linebreakand}{%
  \end{@IEEEauthorhalign}
  \hfill\mbox{}\par
  \mbox{}\hfill\begin{@IEEEauthorhalign}
}
\makeatother

\begin{document}

\title{Hyperbolic Space Learning Method Leveraging Temporal Motion Priors for Human Mesh Recovery}
% Hyperbolic Space Learning Method Leveraging Temporal Motion Priors for Human Mesh Recovery
% Hyperbolic Space Learning Based On Temporal Motion Priors for Human Mesh Recovery

% \author{Anonymous ICME submission}
\author{\normalsize Xiang Zhang, Suping Wu*, Weibin Qiu, Zhaocheng Jin, Sheng Yang
  \\
  \normalsize School of Information Engineering, Ningxia University, Yinchuan, 750021, China
  \\   \normalsize zxiang7996@stu.nxu.edu.cn, pswuu@nxu.edu.cn, 12023131986@stu.nxu.edu.cn, nxujzc@163.com, 18791368334@163.com
  \\   \normalsize Corresponding Author: Suping Wu \quad Email: pswuu@nxu.edu.cn
\thanks{* represents corresponding author. This work is supported by Ningxia Natural Science Foundation Project under Grant (2024AAC02012) and in part by the National Natural Science Foundation of China under Grant (62062056).}
  }
% \author{Anonymous ICME submission}
% \author{\IEEEauthorblockN{1\textsuperscript{st} Xiang Zhang}
% \IEEEauthorblockA{\textit{School of Information Engineering} \\
% \textit{Ningxia University}\\
% YinChuan, China \\
% zxiang7996@stu.nxu.edu.cn}
% \and
% \IEEEauthorblockN{2\textsuperscript{nd} Suping Wu*}
% \IEEEauthorblockA{\textit{School of Information Engineering} \\
% \thanks{* represents corresponding author. This work is supported by Ningxia Natural Science Foundation Project under Grant （2024AAC02012）and in part by the National Natural Science Foundation of China under Grant(62062056).}
% \textit{Ningxia University}\\
% YinChuan, China\\
% pswuu@nxu.edu.cn}
% \and
% \IEEEauthorblockN{3\textsuperscript{rd} Weibin Qiu}
% \IEEEauthorblockA{\textit{School of Information Engineering} \\
% \textit{Ningxia University}\\
% YinChuan, China\\
% 12023131986@stu.nxu.edu.cn}
% \linebreakand 
% \IEEEauthorblockN{4\textsuperscript{th} Zhaocheng Jin}
% \IEEEauthorblockA{\textit{School of Information Engineering} \\
% \textit{Ningxia University}\\
% YinChuan, China\\
% nxujzc@163.com}
% \and
% \IEEEauthorblockN{5\textsuperscript{th} Sheng Yang}
% \IEEEauthorblockA{\textit{School of Information Engineering} \\
% \textit{Ningxia University}\\
% YinChuan, China\\
% 18791368334@163.com}

% }

\maketitle

\begin{abstract}
3D human meshes show a natural hierarchical structure (like torso-limbs-fingers). But existing video-based 3D human mesh recovery methods usually learn mesh features in Euclidean space. It's hard to catch this hierarchical structure accurately. So wrong human meshes are reconstructed. To solve this problem, we propose a hyperbolic space learning method leveraging temporal motion prior for recovering 3D human meshes from videos. First, we design a temporal motion prior extraction module. This module extracts the temporal motion features from the input 3D pose sequences and image feature sequences respectively. Then it combines them into the temporal motion prior. In this way, it can strengthen the ability to express features in the temporal motion dimension. Since data representation in non-Euclidean space has been proved to effectively capture hierarchical relationships in real-world datasets (especially in hyperbolic space), we further design a hyperbolic space optimization learning strategy. This strategy uses the temporal motion prior information to assist learning, and uses 3D pose and pose motion information respectively in the hyperbolic space to optimize and learn the mesh features. Then, we combine the optimized results to get an accurate and smooth human mesh. Besides, to make the optimization learning process of human meshes in hyperbolic space stable and effective, we propose a hyperbolic mesh optimization loss. Extensive experimental results on large publicly available datasets indicate superiority in comparison with most state-of-the-art. 
% 该策略以时序运动先验的信息辅助学习，在双曲空间分别利用3D姿态和姿态运动信息优化学习人体网格特征。
% we further design a hyperbolic pose optimization module and a hyperbolic motion optimization module.
% The former optimizes the initial mesh through the interactive learning between the human mesh and pose. The latter achieves the same purpose through the interactive learning between the human mesh and temporal motion information. Then, we combine the optimized mesh to get the accurate and smooth human mesh. Besides, to make the optimization learning process of human meshes in hyperbolic space stable and effective, we propose a hyperbolic mesh optimization loss. Extensive experimental results on large publicly available datasets indicate superiority in comparison with most state-of-the-art. 

\end{abstract}


\section{Introduction}
\label{sec:intro}
\begin{figure}  
\centering  
\includegraphics[width=0.49\textwidth, height=0.32\textheight]{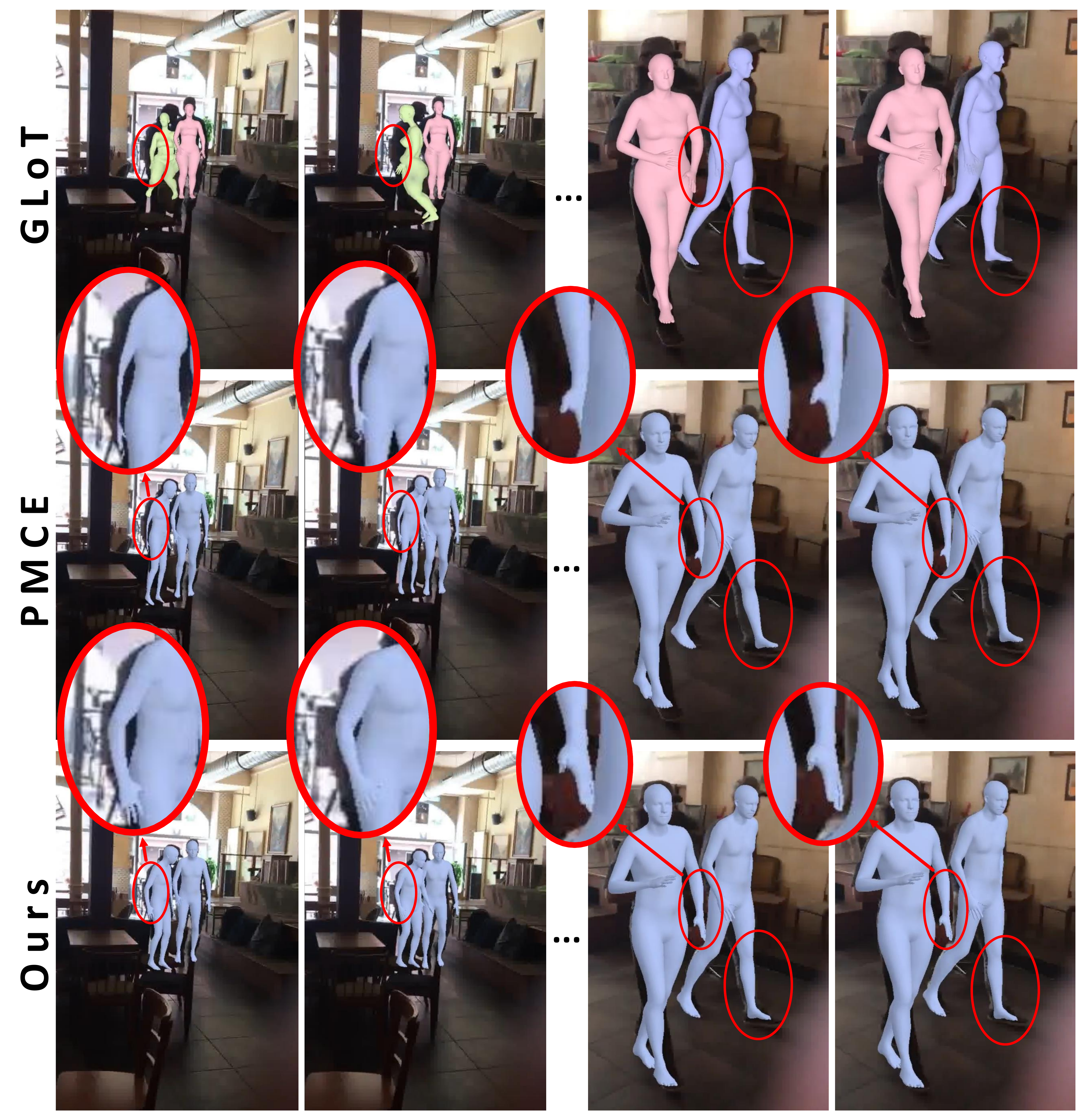}  
\vspace{-0.8cm}\caption{In extreme illumination scenes, our method can recover reasonable human mesh (no limb atrophy or malposition) compared to the SOTA method.}   
\vspace{-0.8cm}  
\label{fig1}  
\end{figure}

3D mesh reconstruction has received wide attention and application in fields like VR, AR and virtual fitting. Although image-based 3D human reconstruction methods have made great progress in accuracy, when applied to video sequences, frame-by-frame reconstruction will make them have serious motion jitter. For this, video-based methods use time information to improve the accuracy and time consistency of human meshes within video frames. For example, TCMR \cite{Choi2020BeyondSF} uses a GRU-based time encoder to predict the current frame based on past and future frames to strengthen time consistency. GLoT \cite{Shen2023GlobaltoLocalMF} separates short-term and long-term time modeling, and uses global and local Transformers for global motion modeling and local parameter correction. Bi-CF \cite{wu2023clip} introduces a two-layer Transformer to model the temporal dependency relationships within video clips and between different clips. PMCE \cite{You2023CoEvolutionOP} presents a two-stream encoder to handle 2D pose sequences and static image features, respectively. And it designs a collaborative progressive decoder to use image-guided adaptive layer normalization to perform pose and mesh interaction. STAF \cite{yao2024staf} uses motion coherence via attention-based time fusion, extracts spatial alignment from predicted mesh projection, and applies multi-level adjacent fusion to enhance feature representation. DGTR \cite{Tang2024DualBranchGT} proposes a dual-branch graph transformer network. It uses a dual-branch with a global motion-attention and a local detail-refinement one to extract long-term and local key info in parallel. DiffMesh \cite{zheng2025diffmesh} effectively generates accurate and smooth output mesh sequences by adding human motion in the forward and reverse processes of the diffusion model. Although these methods have improved performance, they all learn and model 3D human meshes in Euclidean space. In Fig. \ref{fig1}, They don't consider that it is difficult to capture the hierarchical structure (e.g., the positional and shape relationships among the shoulder, arm and hand.) of human meshes in Euclidean space. As a result, the reconstructed meshes are wrong.
% UNSPAT presents an uncertainty-guided Transformer. It can effectively integrate spatial information into the time axis. And it uses an uncertainty-guided attention reweighting module to improve performance.

\begin{figure*}  
\centering  
\includegraphics[width=1.0\textwidth, height=0.25\textheight]{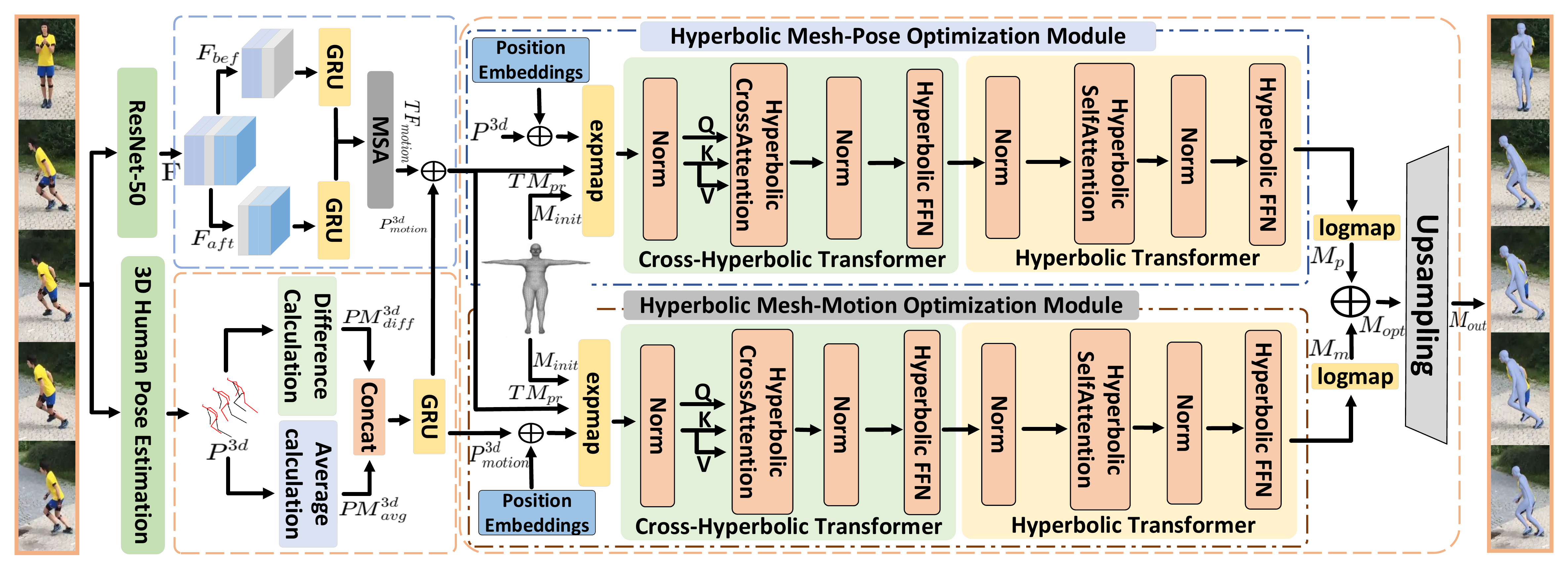}  
\vspace{-0.8cm}\caption{The overall architecture of the our method. We use the pre-trained ResNet-50 and 3D pose estimator to extract $F$ and $P^{3d}$ of the input video $V$. Then, we extract the $TF_{motion}$ and the $P_{motion}^{3d}$ from the $F$ and the $P^{3d}$ respectively, and fuse them into the $TM_{pr}$. We apply expmap to map the input information to the hyperbolic space. With the assistance of the $TM_{pr}$ information, we make the $M_{init}$ interact and learn with $P^{3d}$ and $P_{motion}^{3d}$ respectively to optimize the mesh features. The optimized result $M_{opt}$ is mapped back to the Euclidean space and upsampled to get the accurate human mesh $M_{out}$.}   
\vspace{-0.5cm}  
\label{fig2}  
\end{figure*}

To solve this problem, we propose a hyperbolic space learning method leveraging temporal motion prior for recovering 3D human meshes from videos. First, to fully extract the temporal human motion information in the input 3D pose sequences and image feature sequences, we design a temporal motion prior extraction module. For 3D pose sequences, we do a temporal difference operation and calculate the average motion, combine them for comprehensive information, then use GRU to extract temporal implicit features as temporal pose motion information. For image feature sequences, divide them by the middle frame, use GRU to extract temporal information respectively, apply the attention mechanism to dynamically fuse motion features of different periods and output temporal motion features. Finally, integrate the temporal pose motion information into the temporal motion features to obtain the temporal motion prior, which effectively improves the temporal motion representation ability of the features. 

% When dealing with 3D pose sequences, we first do a temporal difference operation on the 3D pose sequences to accurately catch the detailed changes of joint movements. Meanwhile, we calculate the average motion of poses to show the overall trend. we combine the two to get comprehensive pose motion information. Then, we use GRU \cite{dey2017gate} to extract the temporal implicit features in the pose motion information, and regard them as the temporal pose motion information. When dealing with the image feature sequences, first divide them into two parts, the front part and the back part, according to the middle frame. Then use GRU \cite{dey2017gate} to extract the temporal information respectively, so as to focus on the temporal motion changes in different periods. After that, we use the attention mechanism, which learn the weights according to the feature relationships in different periods, dynamically fuse the motion features of different periods by weighted summation and output the temporal motion features. Finally, we integrate the temporal human motion information into the temporal motion features to get the temporal motion prior. This process can effectively improve the temporal motion representation ability of the features. 

% Data representation in non-Euclidean space has been proved to be able to effectively capture the hierarchical and complex relationships in real-world datasets \cite{FeinAshley2024HVTAC,yang2024hypformer,Glehre2018HyperbolicAN}. 
The hyperbolic space \cite{FeinAshley2024HVTAC,yang2024hypformer,Glehre2018HyperbolicAN} has been proven to be able to effectively capture the hierarchical structures. And 3D human meshes have a natural hierarchical structure. So it's more reasonable to learn and optimize mesh features in hyperbolic space. For this, we design a hyperbolic space optimization learning strategy. By expanding the optimization learning of mesh features to the hyperbolic space, we can model the hierarchical structure characteristics of human meshes more effectively. This strategy uses the temporal motion prior to assist learning. It is divided into two modules: the hyperbolic pose optimization module (HPO) and the hyperbolic motion optimization (HMO) module. The HPO module mainly uses static poses to optimize human mesh learning. The HMO module focuses on using the temporal pose motion information to optimize human mesh learning. This mechanism enables the model to consider both the accuracy of static poses and the continuity of dynamic motions, so as to produce more accurate and smoother human mesh reconstruction results. Our contributions are summarized as follows:

% 非欧几里得空间中的数据表示已被证明可有效捕捉现实世界数据集中的层次化和复杂关系。尤其是双曲空间，可为层次结构提供有效的嵌入。而3D人体网格具有天然的层次结构，因此在双曲空间中对网格特征进行学习和优化更加合理。为此，我们设计了双曲网格优化学习策略，通过将特征学习扩展到双曲空间，更有效地建模人体的层次结构特性。在我们的设计中，采用 PoincareBall 模型作为双曲空间的具体实现，这使得我们能够在保持计算效率的同时，充分利用双曲几何的优势。双曲网格优化学习策略分为两个模块：双曲人体姿态优化模块和双曲人体运动优化模块。双曲人体姿态优化模块主要利用静态姿态优化人体网格学习。双曲人体运动优化模块则着重于利用时序人体运动先验信息优化人体网格学习。这种机制使得模型能够同时考虑静态姿态的准确性和动态运动的连续性，从而产生更加准确且平滑的人体网格重建结果。

\begin{itemize}
    \item We propose a hyperbolic space learning method leveraging temporal motion prior for recovering 3D human meshes from videos, which effectively improves the accuracy and smoothness of the reconstructed human meshes.
    \item We design a temporal motion prior extraction module to fully extract human motion prior information from image feature sequences and 3D pose sequences. Besides, in order to model the hierarchical structure characteristics of the mesh more effectively in hyperbolic space, we also propose a hyperbolic mesh optimization learning strategy. 
    \item As far as we know, we are the first to adopt the method of learning mesh features in the hyperbolic space. Our method's experimental results demonstrate its effectiveness in widely evaluated benchmarks compared to SOTA.
\end{itemize}

% The overall architecture of the hyperbolic space learning method leveraging temporal motion prior proposed by us is shown in Fig. \ref{fig2}.
\section{Method}
The overall architecture of our method is shown in Fig. \ref{fig2}. It mainly consists of two parts: 1) Temporal motion prior extraction. 2) Hyperbolic space optimization learning. Specifically, given a video sequence $V =\lvert\{I_{t}\rvert\}_{t=1}^{T}$ that contains $T$ frames, we use the pre-trained ResNet50 to extract the image features $F\in\mathbb{R}^{T\times2048}$ of each frame. Besides, we use an off-the-shelf 2D pose estimator to estimate the 2D pose $P^{2d}$, and then adopt a spatio-temporal Transformer network \cite{zhang2022mixste} to lift the 2D pose $P^{2d}\in\mathbb{R}^{T\times J\times2}$ to the 3D pose $P^{3d}\in\mathbb{R}^{T\times J\times3}$, where $J$ represents the number of body joints. In the process of extracting the temporal motion prior, on the one hand, we extract the temporal pose motion information $P_{motion}^{3d}\in\mathbb{R}^{T\times J\times3}$ from the 3D pose sequence. On the other hand, we divide the image feature sequence into two parts, $F_{bef}\in\mathbb{R}^{(T/2)\times2048}$ and $F_{aft}\in\mathbb{R}^{(T/2)\times2048}$, from the position of the middle frame. Then we extract the temporal information respectively and obtain the temporal motion feature $TF_{motion}\in\mathbb{R}^{T\times2048}$ through the attention mechanism. After that, we fuse the temporal pose motion information and the temporal motion feature into the temporal motion prior $TM_{pr}\in\mathbb{R}^{T\times2048}$. In the process of hyperbolic space optimization learning, with the help of the temporal motion prior information, we use the 3D pose $P^{3d}$ and the temporal pose motion information $P_{motion}^{3d}$ respectively to optimize the input initial human mesh features $M_{init}\in\mathbb{R}^{431\times3}$ in the hyperbolic space. Then we get the pose-optimized mesh $M_p$ and the motion-optimized mesh $M_m$ and fuse them to obtain a reasonable intermediate human mesh $M_{opt}$. Finally, we upsample the intermediate human mesh to get the fine human mesh $M_{out}\in\mathbb{R}^{6890\times3}$ and output it. We will explain each part in detail in the following sections. 

\subsection{Temporal Motion Prior Extraction}
To fully extract temporal human motion prior from the $F$ and $P^{3d}$, we design a temporal motion prior extraction module.
\paragraph{Temporal Pose Motion Information Extractor}
First, we use the way of calculating the position differences of adjacent joints ${PM}_{diff}^{3d}$ in the time sequence to capture the changes of the joints as time passes. Then, we calculate the average motion ${PM}_{avg}^{3d}$. The ${PM}_{avg}^{3d}$ as a representation of the overall trend of the whole motion sequence, which is combined with the previous differential motion information ${PM}_{diff}^{3d}$ to form a comprehensive motion description ${PM}_{cont}^{3d}$ that includes both microscopic motion details and macroscopic motion trends. For the ${PM}_{cont}^{3d}$, GRU is used to extract their implicit features $P_{motion}^{3d}$ in the time sequence, which provide key temporal pose motion information for subsequent feature fusion. This can be formulated as:

\begin{eqnarray}
\begin{array}{lr}
{PM}_{diff}^{3d}=\lvert\{P_t^{3d}-P_{t-1}^{3d}\rvert\}_{t=2}^T,\\
{PM}_{avg}^{3d}=Average({P}^{3d}),\\
{PM}_{cont}^{3d}=Concat[{PM}_{diff}^{3d},{PM}_{avg}^{3d}],\\
P_{motion}^{3d}=GRU({PM}_{cont}^{3d}).
\end{array}
\end{eqnarray}
where, $Average$ represents the calculation of the mean, and $Concat$ is the operation of concatenation.

\paragraph{Temporal Motion Feature Extraction}
For image feature sequences, we divide them into the front part $F_{bef}$ and the back part $F_{aft}$, according to the middle frame, and use GRU to extract the temporal information of these two parts respectively. This division enables the model to focus on the temporal changes of image features in different time periods. Then, we concatenate the front and back parts and input them into the attention mechanism, learn weights according to the feature relationships in different periods, and dynamically fuse the motion features of different periods by weighted summation to highlight the key temporal information. Finally, we add the $P_{motion}^{3d}$ to the result output by the attention mechanism. This helps to incorporate the pose motion information. Also, it effectively improves the ability of the features to represent temporal motion. This can be formulated as:

\begin{eqnarray}
\begin{array}{lr}
F_{bef},F_{aft}=Seg(\lvert\{{F_t}\rvert\}_{t=1}^T),\\
TF_{bef}=GRU(F_{bef}),TF_{aft}=GRU(F_{aft}),\\
TF_{cont}=Concat[TF_{bef},TF_{aft}],\\
TM_{pr}=MSA(TF_{cont})+P_{motion}^{3d}.
\end{array}
\end{eqnarray}
where, $Seg$ is used to divide the front and back frames. $TF_{bef}$ and $TF_{aft}$ are the temporal features of the front and back frames respectively, and $MSA$ is the attention mechanism. 

\subsection{Hyperbolic Space Preliminaries}
The hyperbolic space is a kind of non-Euclidean space. Its negative curvature characteristic makes it especially suitable for representing hierarchical structure data. In our design, we adopt the PoincareBall model \cite{FeinAshley2024HVTAC} as the specific implementation of the hyperbolic space. This enables us to make full use of the advantages of hyperbolic geometry while maintaining computational efficiency. 
\paragraph{PoincareBall Model}
The n-dimensional PoincareBall model is defined as a manifold:

\begin{eqnarray}
\begin{array}{lr}
\mathbb{D}^n=\lvert\{x\in\mathbb{R}^n:\parallel \bm {x}\parallel<1\rvert\}.
\end{array}
\end{eqnarray}
Here, $\lvert\parallel\cdot\rvert\parallel$ represents the Euclidean norm. The Riemannian metric tensor $g_x$ of this manifold is as follows: 

\begin{eqnarray}
\begin{array}{lr}
g_{\bm{x}} = \lambda_{\bm{x}}^2 g^E, \quad \lambda_{\bm{x}} = \frac{2}{1 - \| \bm{x} \|^2}.
\end{array}
\end{eqnarray}
Here, $g^E$ is the Euclidean metric tensor, and $\lambda_{\bm{x}}$ is the conformal factor that scales the Euclidean metric to take into account the curvature of the hyperbolic space.

\paragraph{Hyperbolic Lineal Layer}
To apply the Transformer architecture to optimize human mesh features in the hyperbolic space, inspired by \cite{FeinAshley2024HVTAC,yang2024hypformer,Glehre2018HyperbolicAN}, we modify the key components to operate in the hyperbolic space. First, we construct a linear layer adapted to the hyperbolic space by using Möbius matrix-vector multiplication and addition. The Möbius transformation is very important for handling vectors in the PoincareBall model:

\begin{eqnarray}
\begin{array}{lr}
\mathbf{x}\oplus\mathbf{y}=\frac{(1+2\langle\mathbf{x},\mathbf{y}\rangle+\|\mathbf{y}\|^2)\mathbf{x}+(1-\|\mathbf{x}\|^2)\mathbf{y}}{1+2\langle\mathbf{x},\mathbf{y}\rangle+\|\mathbf{x}\|^2\|\mathbf{y}\|^2},\\
\mathbf{W}\otimes_M\mathbf{x}=\tanh\left(\left|\frac{\|\mathbf{Wx}\|}{\|\mathbf{x}\|}\tanh^{-1}(\|\mathbf{x}\|)\right|\right)\frac{\mathbf{Wx}}{\|\mathbf{Wx}\|},\\
h = \mathbf{W} \otimes_M \mathbf{x} \oplus \mathbf{b}.
\end{array}
\end{eqnarray}
Here, $\oplus$ and $\otimes_M$ are Möbius addition and matrix-vector multiplication respectively. $\oplus$ and $\otimes_M$ respectively extend vector addition and matrix-vector multiplication to the hyperbolic space. We adapt the traditional linear layer to the hyperbolic space by using $\otimes_M$ and then adding a bias. 

\begin{eqnarray}
\begin{array}{lr}
h=W\otimes_Mx\oplus b.
\end{array}
\end{eqnarray}

where, $W$ is a matrix and $b$ is a bias. This construction enables us to perform linear transformations in the hyperbolic space.

\paragraph{Hyperbolic Activation Functions and Normalization}
We apply the exponential map (expmap0) to map features from the Euclidean space to the hyperbolic space:

\begin{eqnarray}
\begin{array}{lr}
\exp_0(\mathbf{v})=\tanh\lvert(\frac{\parallel\mathbf{v}\parallel}{2}\rvert)\frac{\mathbf{v}}{\parallel\mathbf{v}\parallel}.
\end{array}
\end{eqnarray}
The logarithmic map (logmap0) brings the features back from the hyperbolic space to the Euclidean space:

\begin{eqnarray}
\begin{array}{lr}
\log_0(x)=\frac{2\tanh^{-1}(\parallel x\parallel)}{\parallel x\parallel}x.
\end{array}
\end{eqnarray}
Using these mappings, we define the hyperbolic space versions of the activation function GELU and AdaLayerNorm \cite{huang2017arbitrary}:

\begin{eqnarray}
\begin{array}{lr}
\widehat{GELU}(\mathbf{x})=\exp_0\lvert(GELU\lvert(\log_0(\mathbf{x})\rvert)\rvert),\\
\widehat{AdaLN}(\mathbf{x})=\exp_0\lvert(AdaLN\lvert(\log_0(\mathbf{x})\rvert)\rvert).
\end{array}
\end{eqnarray}

\subsection{Hyperbolic Space Optimization Learning}

In the task of 3D human mesh reconstruction, the human mesh structure has a natural hierarchical relationship (such as torso-limbs-fingers). Traditional Euclidean space methods often have limitations in expressive ability when dealing with such hierarchical structure data. For this reason, we propose a hyperbolic space optimization learning strategy. By expanding the optimization learning of human mesh features to the hyperbolic space, we can model the hierarchical structure of the human mesh more effectively. 
% First, we use the exponential map (expmap0) to transform the input information into the hyperbolic space (including 3D human poses $P^{3d}$, pose motion information $P_{motion}^{3d}$, initial mesh $M_{init}$ and temporal motion prior $TM_{pr}$). 

% \begin{eqnarray}
% \begin{array}{lr}
% {\hat{M}}_{init}=\exp_0\funcapply\lvert(M_{init}\rvert),
% {\widehat{TM}}_{pr}=\exp_0\funcapply\lvert(TM_{pr}\rvert),\\
% {\hat{P}}^{3d}=\exp_0\funcapply\lvert(P^{3d}\rvert),
% {\hat{P}}_{motion}^{3d}=\exp_0\funcapply\lvert(P_{motion}^{3d}\rvert).
% \end{array}
% \end{eqnarray}
We use the hyperbolic pose optimization module ($HPO$) and the hyperbolic motion optimization module ($HMO$) respectively to optimize the initial mesh. We fuse the mesh features output by these two modules into the intermediate human mesh $M_{opt}$, and upsample the intermediate human mesh to get the fine human mesh $M_{out}$ and output it. This can be formulated as:

\begin{eqnarray}
\begin{array}{lr}
M_p=HPO({{M}}_{init},{{TM}}_{pr},{{P}}^{3d})\\
M_m=HMO({{M}}_{init},{{TM}}_{pr},{{P}}_{motion}^{3d})\\
M_{opt}=M_p+M_m,\\
M_{out} = Upsampling(M_{opt}).
\end{array}
\end{eqnarray}
% The module designs a hyperbolic space cross-attention mechanism to enable the features of joints and vertices to interact and learn effectively in the hyperbolic space. 
\paragraph{Hyperbolic Pose Optimization Module}
The hyperbolic pose optimization module mainly uses static pose to optimize human mesh learning. Firstly, this module adds learnable position encoding to keep the spatial information. Then, it transforms the $P^{3d}$, the $TM_{pr}$ and the $M_{init}$ into the hyperbolic space by using the exponential map ($\exp_0$), and conducts feature learning and optimization in this space. We use hyperbolic adaptive normalization layers ($\widehat{AdaLN}$) that injects shape and temporal motion information from the $TM_{pr}$ into mesh features while preserving spatial structure. The module designs a hyperbolic cross-attention to enable the features of joints and vertices to interact and learn effectively in the hyperbolic space. This can be formulated as:
\begin{eqnarray}
\begin{array}{lr}
{\hat{M}}_{init}=\exp_0(M_{init}),
{\widehat{TM}}_{pr}=\exp_0(TM_{pr}),\\
{\hat{P}}^{3d}=\exp_0(P^{3d}),
{\hat{P}}_{motion}^{3d}=\exp_0(P_{motion}^{3d}), \\
{\hat{M}}_{mix}=\widehat{AdaLN}({\hat{M}}_{init},{\widehat{TM}}_{pr}),\\
Q_M={\hat{M}}_{mix}\otimes_MW_Q, \\K_P={\hat{P}}^{3d}\otimes_MW_K,V_P={\hat{P}}^{3d}\otimes_MW_V,\\
{\hat{X}}_{PM}=HYMCA(Q_M,K_P,V_P)\oplus{\hat{M}}_{mix},\\
{\hat{X}}_{ada} = \widehat{AdaLN}({\hat{X}}_{PM},{\widehat{TM}}_{pr}),\\
{\hat{X}}_M=HYFFN({\hat{X}}_{ada})\oplus{\hat{X}}_{PM}.
\end{array}
\end{eqnarray}
 Meanwhile, we use the hyperbolic self-attention to enhance the feature extraction ability. Finally, it maps ($\log_0$) the optimized mesh features $\hat{M_p}$ back to the Euclidean space.
 % using the logarithmic map ($\log_0$)

\begin{eqnarray}
\begin{array}{lr}
Q_M^\prime={\hat{X}}_M\otimes_MW_Q,\\K_P^\prime={\hat{X}}_M\otimes_MW_K, V_P^\prime={\hat{X}}_M\otimes_MW_V,\\
\hat{X_p}=HYMSA\lvert(Q_M^\prime,K_P^\prime,V_P^\prime\rvert)\oplus{\hat{X}}_M,\\
\hat{M_p}=HYFFN(\widehat{AdaLN}(\hat{X_p},{\widehat{TM}}_{pr})\oplus\hat{X_p},\\
M_p=\log_0(\hat{M_p}).
\end{array}
\end{eqnarray}
here, $W_{(\cdot)}$ is the weighting matrix. $HYMCA$ is hyperbolic cross-attention, $HYMSA$ is hyperbolic self-attention, and $HYFFN$ is hyperbolic multi-layer perceptron. The linear layers and activation functions in them all adopt adjusted forms that are suitable for the hyperbolic space.
% $P_{motion}^{3d}$ and the $M_{init}$
\paragraph{Hyperbolic Motion Optimization Module}
The hyperbolic motion optimization module focuses on using the temporal pose motion information $P_{motion}^{3d}$ to optimize human mesh learning. Firstly, this module adds learnable position encoding to enhance the spatial perception ability. Then, it transforms the $P_{motion}^{3d}$, $TM_{pr}$ and the $M_{init}$ into the hyperbolic space by using the $\exp_0$ for optimization learning. In hyperbolic space, the module uses the hyperbolic cross-attention to achieve the interaction learning of pose motion information and mesh features, so that the mesh features can sense and learn the hierarchical temporal motion information contained in the $P_{motion}^{3d}$. Next, the module uses the hyperbolic self-attention to further enhance the expression ability of the mesh features, enabling them to better capture the deformation patterns in the motion process. Finally, it maps the optimized mesh features $\hat{M_m}$ back to the Euclidean space by using the $\log_0$. The specific formula is similar to the Formula 11 and 12, and only the inputs are different. This feature interaction and optimization strategy based on the hyperbolic space enables the module to effectively inject the motion prior information into the mesh feature learning and improves the accuracy and smoothness of the reconstructed mesh. 
% \paragraph{双曲人体运动优化模块}

\subsection{Loss Function}
Due to the non-Euclidean characteristics of the hyperbolic space, it's challenging to optimize our model in the hyperbolic space. To ensure the stability and effectiveness of our model training process, we design a hyperbolic mesh optimization loss. In this loss, we first transform the 3D mesh ground truth and the predicted value into the hyperbolic space, and then calculate the difference between them by using the L1 loss.

\begin{eqnarray}
\begin{array}{lr}
{\hat{M}}_{gt}=\exp_0\lvert(M_{gt}\rvert),
{\hat{M}}_{pre}=\exp_0\lvert(M_{pre}\rvert),\\
\mathcal{L}_{hymesh}=\frac{1}{V}\sum_{i=1}^{V}\lvert\|{\hat{M}}_{gt}-{\hat{M}}_{pre}\rvert\|_{1}.
\end{array}
\end{eqnarray}
Following \cite{You2023CoEvolutionOP,Choi2020Pose2MeshGC}, we employ the mesh vertex loss $\mathcal{L}_{mesh}$, 3D joint loss $\mathcal{L}_{joint}$, surface normal loss $\mathcal{L}_{normal}$, surface edge loss $\mathcal{L}_{edge}$ and hyperbolic mesh optimization loss $\mathcal{L}_{hymesh}$ to constrain the parameter learning of the network.
\begin{eqnarray}
\begin{array}{lr}
\mathcal{L_{E}}=\lambda_m\mathcal{L}_{mesh}+\lambda_j\mathcal{L}_{joint}+\lambda_n\mathcal{L}_{normal}+\lambda_e\mathcal{L}_{edge},\\
\mathcal{L}=\mathcal{L_{E}}+\lambda_{hy}\mathcal{L}_{hymesh}.
\end{array}
\end{eqnarray}
Where $\lambda_m=\lambda_j=\lambda_{hy}=1$, $\lambda_n=0.1$, and $\lambda_e=20$. $\mathcal{L_{E}}$ is the Euclidean loss, and $\mathcal{L}_{hymesh}$ is the hyperbolic loss.

\begin{table*}[htbp] 
  \centering
  \caption{Evaluation of SOTA methods on 3DPW, Human3.6M and MPI-INF-3DHP datasets. All methods use pre-trained ResNet50 to extract the features of each frame of the image.}
  \vspace{-0.3cm} 
    \scalebox{0.7}{\begin{tabular}{c|cccc|ccc|ccc|c}
    \bottomrule
    \multirow{2}[4]{*}{Method} & \multicolumn {4}{c|}{3DPW}   & \multicolumn{3}{c|}{Human3.6M} & \multicolumn{3}{c|}{MPI-INF-3DHP} &\multirow{2}[4]{*}{\shortstack{numbe of \\ input frame}} \\
\cmidrule{2-11}         & MPJPE$\downarrow$ & PA-MPJPE$\downarrow$  & MPVPE$\downarrow$ & ACC-ERR$\downarrow$ & MPJPE$\downarrow$ & PA-MPJPE$\downarrow$  & ACC-ERR$\downarrow$ & MPJPE$\downarrow$ & PA-MPJPE$\downarrow$  & ACC-ERR$\downarrow$ &  \\
    \midrule
    % MPS-Net\cite{Wei2022CapturingHI}  & 84.3  & 52.1   & 99.7    & 7.4     & 69.4    & 47.4   & 3.6 & 96.7    & 62.8   & 9.6  &16  \\
    % GLAMR (CVPR 22)\cite{yuan2022glamr}  & -  & 51.1   & -    & 8.9     & -    & 47.6   & 6.0 & - & - & - &16   \\
    % Zhang et al. (CVPR 23)\cite{zhang2023two}  & 83.4  & 51.7   & 98.9    & 7.2     & 73.2    & 51   & 3.6 & 98.2    & 62.5   & 8.6  &16  \\
    GloT (CVPR 23)\cite{Shen2023GlobaltoLocalMF}  & 84.3  & 50.6   & 96.3    & 6.6     & 67.0    & 46.3   & 3.6 & 93.9    & 61.5   & 7.9  &16  \\
    Bi-CF (ACM MM 23)\cite{wu2023clip}  & 73.4  & 51.9   & 89.8    & 8.8     & 63.9    & 46.1   & \bfseries3.1 & 95.5    & 62.7   & 7.7  &16  \\
    Key2Mesh (CVPR 24)\cite{uguz2024mocap} & 86.7  & 49.8   & 99.5    & -     & 107.1    & 51.0   & 3.4 & - & - & - & 16  \\
    STAF (TCSVT 24)\cite{yao2024staf} & 80.6  & 48.0   & 95.3    & 8.2     & 70.4    & 44.5   & 4.8  & 93.7    & 59.6   & 10.0  & 16\\
    
    PMCE (ICCV 23)\cite{You2023CoEvolutionOP}  & \underline{69.5}  & \underline{46.7}   & \underline{84.8}    & 6.5     & \underline{53.5}    & \underline{37.7}   & \bfseries3.1 & 79.7    & 54.5   & \underline{7.1}  &16  \\
    DGTR (IROS 24)\cite{Tang2024DualBranchGT} & 82.0  & 51.3   & 97.3    & 7.6     & 67.2    & 46.1   & 3.8  & 94.5    & 61.3   & 8.5  & 16\\
    % UNSPAT (WACV 24)\cite{lee2024unspat} & 75.0  & \bfseries45.5   & 90.2    & 6.5     & 58.3    & 41.3   & 3.8  & 94.4    & 60.4   & 9.2  & 16\\
    DiffMesh (WACV 25)\cite{zheng2025diffmesh} & 77.2  & 48.5   & 94.0    & \bfseries{6.3}     & 65.3    & 41.9   & \underline{3.3}  & \underline{78.9}   & \underline{54.4}   & \bfseries7.0  & 16\\
    Ours  & \bfseries68.2  & \bfseries 46.4  & \bfseries80.7      &\underline{6.4}     &\bfseries51.4     & \bfseries36.7     &\bfseries 3.1  &\bfseries73.0    &\bfseries 53.1   & \underline{7.1}  &16  \\
    \bottomrule
    \end{tabular}}\vspace{-0.5cm}
  \label{tab1}%
\end{table*}%

\section{Experiments}
\subsection{Implementation Details}
According to \cite{You2023CoEvolutionOP} and \cite{Choi2020Pose2MeshGC}, we set the input sequence length to 16 and use the pre-trained ResNet-50 as the image feature extractor. For 2D pose detection and 3D pose estimation, we adopt the same configuration as described in \cite{You2023CoEvolutionOP} and \cite{Choi2020Pose2MeshGC}. Both training stages of the network are optimized using Adam. We train the entire network with a batch size of 32 and a learning rate of $5\times{10}^{-5}$ for 10 epochs. Our code is implemented in Pytorch, and the training is performed on an NVIDIA RTX 4090 GPU.
% from SPIN\cite{Kolotouros2019LearningTR}
\subsection{Evaluation Datasets and Metrics}
Building upon previous approaches\cite{You2023CoEvolutionOP} and \cite{wei2022capturing}, we utilize a combination of 2D and 3D datasets for training our model. In particular, we utilize the 3D datasets including 3DPW\cite{von2018recovering}, Human3.6M\cite{ionescu2013human3}, and MPI-INF-3DHP\cite{mehta2017monocular} with annotations for 3D joints and SMPL parameters. For 2D datasets, we incorporate COCO\cite{lin2014microsoft} and MPII\cite{andriluka20142d}. We calculated the mean error per joint position (MPJPE) and Procrustesaligned MPJPE (PA-MPJPE) as the main metrics of accuracy. And we measured the Euclidean distance (MPVPE) between the ground truth vertex and the predicted vertex. In addition, we calculated the mean of the difference between the predicted 3D coordinates and the ground truth acceleration (Accel) for the temporal evaluation.
% \cite{Loper2023SMPLAS}

\begin{table}[htbp]\vspace{-0.5cm}
  \centering
  \caption{Performance comparison with SOTA video-based methods on 3DPW without using 3DPW training set during training.}
  \vspace{-0.3cm} 
    \scalebox{0.65}{\setlength{\tabcolsep}{1.0mm}\begin{tabular}{c|cccc}
    \toprule
    Mesh Recovery (w/o 3DPW in Train) & \multicolumn{1}{l}{MPJPE$\downarrow$} & \multicolumn{1}{l}{PA-MPJPE$\downarrow$} & \multicolumn{1}{l}{MPVPE$\downarrow$} & \multicolumn{1}{l}{ACC-ERR$\downarrow$}\\
    \midrule
    % PQ-GCN\cite{wang2022progressive} & 89.2  & 58.3 & 106.4 & - \\
    Pose2Mesh\cite{Choi2020Pose2MeshGC} & 88.9 & 58.3 & 106.3 & 22.6 \\
    % GTRS\cite{zheng2022lightweight} & 88.5  & 58.9 & 103.2 & 25 \\
    GloT\cite{Shen2023GlobaltoLocalMF} & 89.9  & 53.5 & 107.8 & 6.7 \\
    PMCE\cite{You2023CoEvolutionOP} & 81.6  & \underline{52.3} & 99.5 & 6.8 \\
    Bi-CF\cite{wu2023clip} & \underline{78.3}  & 53.7 & \underline{95.6} & 8.6 \\
    DiffMesh\cite{zheng2025diffmesh} & 88.7  & 53.0 & 105.9 & \underline{6.5} \\
    % STAF\cite{yao2024staf} & 81.2  & \bfseries48.7 & 96.0 & 8.2 \\
    Ours & \bfseries75.0 & \bfseries{51.7} & \bfseries93.1 & \bfseries6.2 \\
    \bottomrule
    \end{tabular}}\vspace{-0.5cm}
  \label{tab2}%
\end{table}%

\subsection{Comparison Result and Ablation Study}
\paragraph{Quantitative Results}
In Table \ref{tab1}, we conduct extensive experiments to compare our method with previous video-based methods. The results show that our method outperforms state-of-the-art methods in almost all metrics, achieving the best reconstruction accuracy and competitive motion smoothness. Specifically, compared to the previous state-of-the-art method PMCE, our model achieves a reduction of 1.3\% (from 69.5$mm$ to 68.2$mm$), 6.7\% (from 79.7$mm$ to 73.0$mm$), and 2.1\% (from 53.5$mm$ to 51.4$mm$) in MPJPE metric on 3DPW, MPI-INF-3DHP, and Human3.6M datasets, respectively. This shows that our strategy of using the temporal motion prior information and optimizing human meshes in the hyperbolic space is reasonable and effective. Despite DiffMesh achieving a marginally better Accel than ours by 0.1 $mm/s^{2}$ on 3DPW and MPI-INF-3DHP dataset, its estimation error MPJPE is significantly higher than ours by 9.0$mm$ and 5.9$mm$. In addition, we report the results on the 3DPW without using the 3DPW training set in Table \ref{tab2}. Our method consistently surpasses those methods across all evaluation metrics. This proves that our method has good generalization ability.

\begin{figure}  
\centering  
\includegraphics[width=0.49\textwidth, height=0.33\textheight]{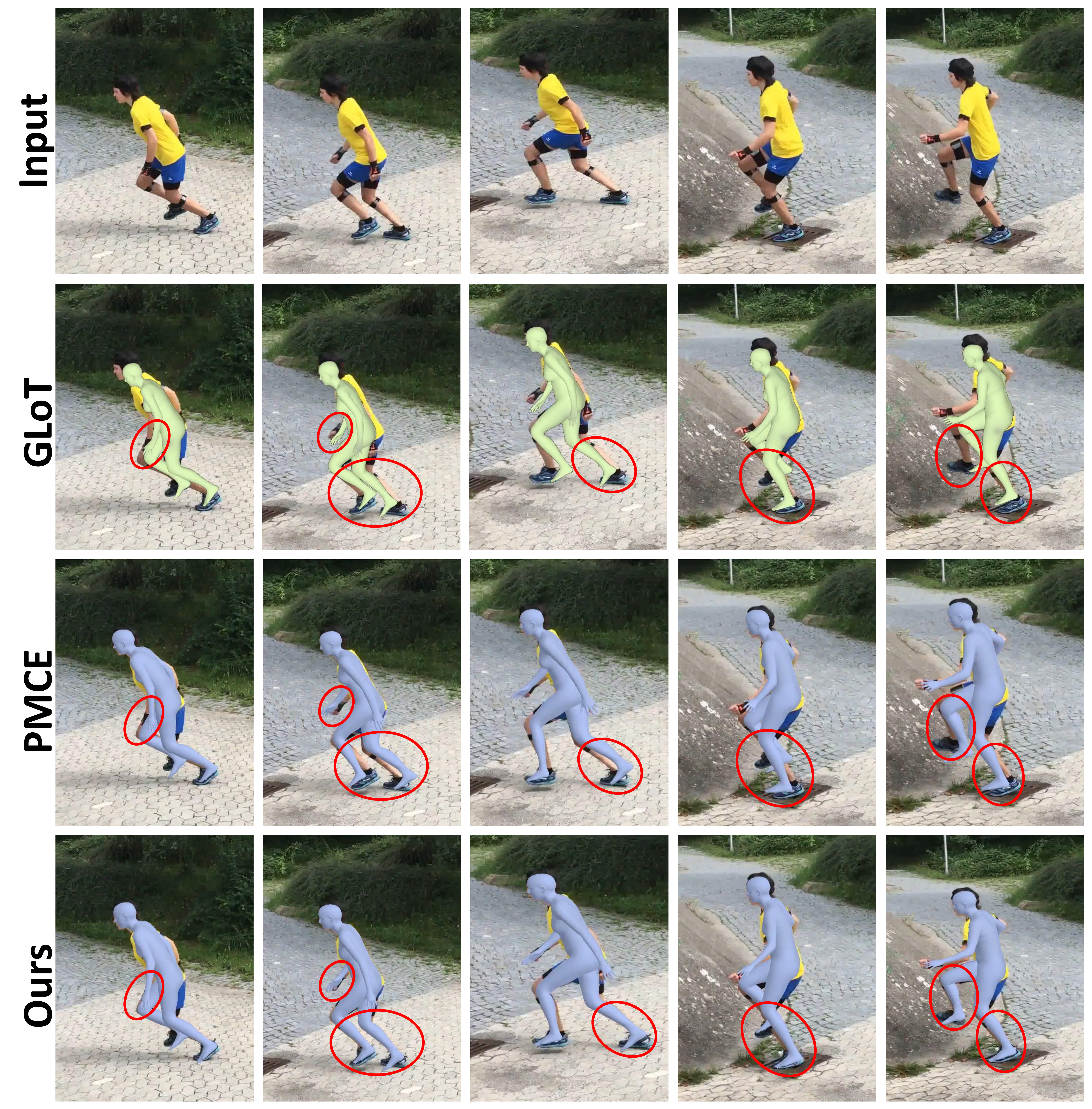} 
\vspace{-0.7cm}\caption{In outdoor fast-motion scenes, our method can recover accurate human mesh compared to the SOTA method.}   
\vspace{-0.5cm}  
\label{fig3}  
\end{figure}
\begin{figure}  
\centering  
\includegraphics[width=0.49\textwidth, height=0.09\textheight]{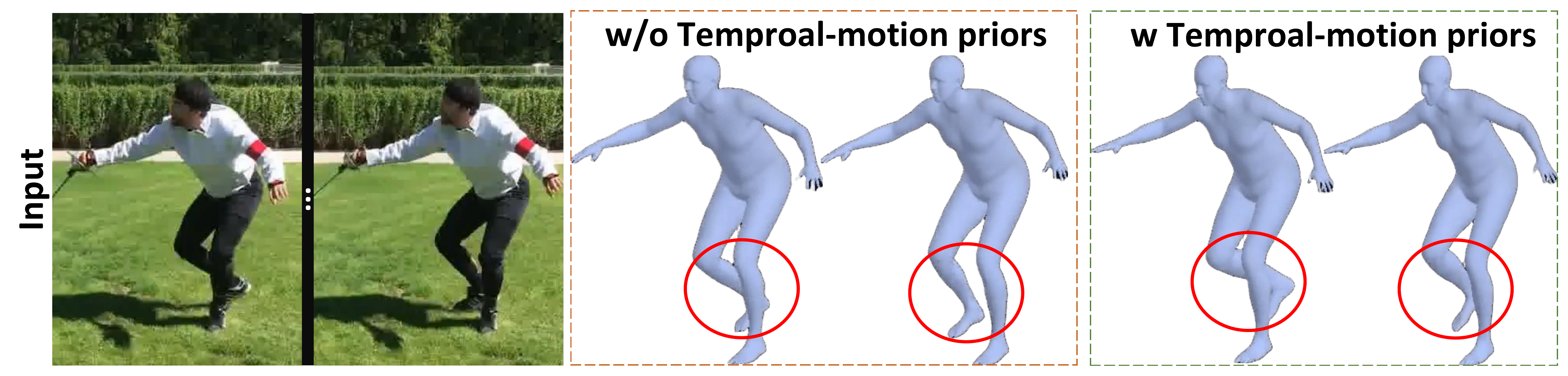} 
\vspace{-0.8cm}\caption{Visual ablation experiments on whether to extract temproal motion priors.}   
\vspace{-0.8cm}  
\label{fig4}  
\end{figure}
\paragraph{Qualitative Results}
Fig. \ref{fig1} shows the qualitative results of our method in extreme illumination scenes. It can be seen that the reconstruction results of GLoT and PMCE are interfered by extreme illumination scenes. GLoT has difficulty reconstructing reasonable human body meshes. Although PMCE can reconstruct the overall human body meshes, because it optimizes and learns the features of human body meshes in the Euclidean space and it's hard to accurately capture the hierarchical structure information of the human body, the reconstructed human bodies have problems like limb atrophy or malposition. Fig. \ref{fig3} shows the qualitative comparison in fast motion scenes. Because of the lack of extraction and utilization of temporal motion priors, it's difficult for the human body meshes reconstructed by GLoT and PMCE to be aligned with the input images. Our method not only fully extracts the temporal motion priors, but also uses pose and motion information respectively in the hyperbolic space to optimize and learn the features of human body meshes. So the reconstructed human body meshes have reasonable movements and are accurately aligned.

\begin{table}[htbp]\vspace{-0.3cm}
  \centering
  \caption{Ablation study on different configurations of our method.}
  \vspace{-0.3cm}
    \scalebox{0.65}{\setlength{\tabcolsep}{1.0mm}\begin{tabular}{ccccc|ccc}
    \toprule
    TPMI-Extractor &TMF-Extractor &HPO &HMO &Hymesh-Loss & \multicolumn{1}{l}{MPJPE$\downarrow$} & \multicolumn{1}{l}{PA-MPJPE$\downarrow$} & \multicolumn{1}{l}{ACC-ERR$\downarrow$} \\
    \midrule
     & & & & & 81.6  & 52.3 & 6.8 \\
    $\checkmark$ & & & & & 81.1  & 52.1 & 6.6 \\
     & $\checkmark$ & & & & 80.2  & 52.2 & 6.4 \\
    $\checkmark$ &$\checkmark$ & & & & 78.5  & 52.1 & 6.3 \\
     & &$\checkmark$ & & & 79.2  & 52.1 & 6.7 \\
     & & & $\checkmark$ & & 79.7  & 52.2 & 6.5 \\
     & & $\checkmark$ & $\checkmark$ & & 77.4  & 51.9 & 6.5 \\
    $\checkmark$ & $\checkmark$ &$\checkmark$ &$\checkmark$ & & 75.7  & 51.9 & 6.2 \\
    % Ours w/o Frequency Spectral Correlation Loss & 26.1  & 20.7  \\
    \midrule
    $\checkmark$ &$\checkmark$ &$\checkmark$ &$\checkmark$ &$\checkmark$  & \bfseries75.0  & \bfseries51.7 & \bfseries6.2 \\
    \bottomrule
    \end{tabular}}\vspace{-0.3cm}
  \label{tab3}%
\end{table}%

\begin{figure}
\centering  
\includegraphics[width=0.49\textwidth, height=0.11\textheight]{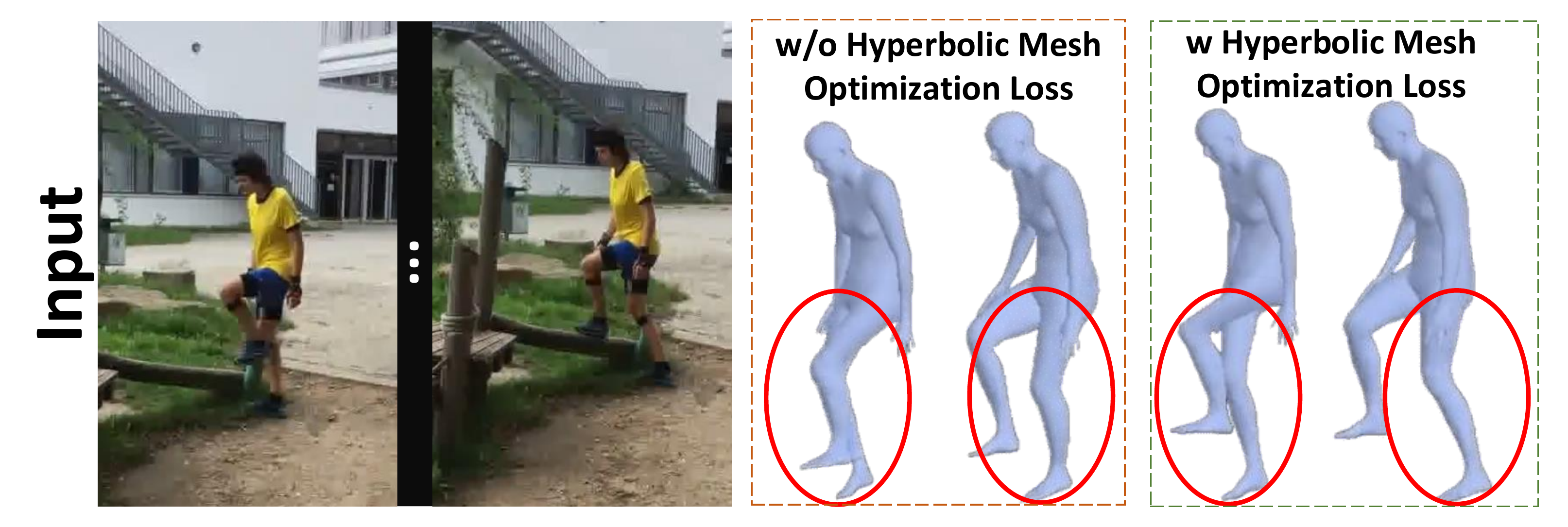} 
\vspace{-0.8cm}\caption{Visual ablation experiments on whether to use Hymesh Loss.}   
\vspace{-0.6cm}  
\label{fig5}  
\end{figure}
\paragraph{Ablation Study}
In Table \ref{tab3}, ablation studies were conducted on different configurations in our method. In the temporal motion prior extractor (TMP-Extractor), using only the Temporal Pose Motion Information Extractor (TPMI-Extractor) or the Temporal Motion Feature Extractor (TMF-Extractor) can't fully extract the temporal motion information, so the improvement of the model's accuracy and smoothness is limited. When we apply these two modules at the same time to extract the prior of human body motion, the improvement of the model's accuracy and smoothness is more obvious. In Fig. \ref{fig4}, the temporal motion prior fully extracted by our designed TMP-Extractor is helpful for the model to reconstruct human body meshes with accurate poses. 
For the hyperbolic space optimization learning strategy (HSOL-Strategy), we conduct ablation tests on the HPO module and the HMO module respectively. The experimental results show that using only pose or motion information to optimize and learn human mesh features can improve the effect, but it's not as significant as using both modules together to optimize the mesh features. Using the TMP-Extractor together with the HSOL-Strategy is helpful to make up for the lack of accuracy when only using the TMP-Extractor and the lack of smoothness when only using the HSOL-Strategy. Finally, we add the hyperbolic mesh optimization loss in the process of training the model to further improve the accuracy of the reconstructed mesh. In Fig. \ref{fig5}, the qualitative ablation experiment on the hyperbolic mesh optimization loss shows that constraining the learning process of our model from the hyperbolic space is crucial for maintaining reasonable poses.

% \paragraph{Qualitative Results in Generalization.}

\section{Conclusion}
We propose a hyperbolic space learning method leveraging temporal motion prior for recovering 3D human meshes from videos. We fully extract the prior information of human motion from the input image feature sequence and 3D human pose sequence to enhance the expression ability of features in the temporal motion dimension. Subsequently, we design a hyperbolic space optimization learning strategy and use 3D pose and pose motion information in the hyperbolic space to optimize and learn the features of human meshes. We also design a hyperbolic mesh optimization loss to ensure the effectiveness and stability of model training. 
% Our approach improves the accuracy of multiple challenge datasets.

% \apptocmd{\thebibliography}{\fontsize{6.2pt}{11pt}\selectfont}{}{}
\bibliographystyle{IEEEbib}
\bibliography{arxiv/arxiv}

\begin{thebibliography}{10}

\bibitem{Choi2020BeyondSF}
Hongsuk Choi, Gyeongsik Moon, et~al.,
\newblock ``Beyond static features for temporally consistent 3d human pose and shape from a video,''
\newblock {\em 2021 IEEE/CVF Conference on Computer Vision and Pattern Recognition (CVPR)}, pp. 1964--1973, 2020.

\bibitem{Shen2023GlobaltoLocalMF}
Xi~Shen, Zongxin Yang, et~al.,
\newblock ``Global-to-local modeling for video-based 3d human pose and shape estimation,''
\newblock {\em 2023 IEEE/CVF CVPR}, pp. 8887--8896, 2023.

\bibitem{wu2023clip}
Peng Wu, Xiankai Lu, et~al.,
\newblock ``Clip fusion with bi-level optimization for human mesh reconstruction from monocular videos,''
\newblock in {\em Proceedings of the 31st ACM MM}, 2023, pp. 105--115.

\bibitem{You2023CoEvolutionOP}
Yingxuan You, Hong Liu, et~al.,
\newblock ``Co-evolution of pose and mesh for 3d human body estimation from video,''
\newblock {\em 2023 IEEE/CVF International Conference on Computer Vision (ICCV)}, pp. 14917--14927, 2023.

\bibitem{yao2024staf}
Wei Yao, Hongwen Zhang, et~al.,
\newblock ``Staf: 3d human mesh recovery from video with spatio-temporal alignment fusion,''
\newblock {\em IEEE Transactions on Circuits and Systems for Video Technology}, 2024.

\bibitem{Tang2024DualBranchGT}
Tao Tang, Hong Liu, et~al.,
\newblock ``Dual-branch graph transformer network for 3d human mesh reconstruction from video,''
\newblock 2024.

\bibitem{zheng2025diffmesh}
Ce~Zheng, Xianpeng Liu, et~al.,
\newblock ``Diffmesh: A motion-aware diffusion framework for human mesh recovery from videos,''
\newblock in {\em IEEE/CVF Winter Conference on Applications of Computer Vision (WACV)}, 2025.

\bibitem{FeinAshley2024HVTAC}
Jacob Fein-Ashley, Ethan Feng, et~al.,
\newblock ``Hvt: A comprehensive vision framework for learning in non-euclidean space,''
\newblock {\em ArXiv}, vol. abs/2409.16897, 2024.

\bibitem{yang2024hypformer}
Menglin Yang, Harshit Verma, et~al.,
\newblock ``Hypformer: Exploring efficient transformer fully in hyperbolic space,''
\newblock in {\em Proceedings of the 30th ACM SIGKDD Conference on Knowledge Discovery and Data Mining}, 2024, pp. 3770--3781.

\bibitem{Glehre2018HyperbolicAN}
Çaglar G{\"u}lçehre, Misha Denil, Mateusz Malinowski, et~al.,
\newblock ``Hyperbolic attention networks,''
\newblock {\em ArXiv}, vol. abs/1805.09786, 2018.

\bibitem{zhang2022mixste}
Jinlu Zhang, Zhigang Tu, et~al.,
\newblock ``Mixste: Seq2seq mixed spatio-temporal encoder for 3d human pose estimation in video,''
\newblock in {\em Proceedings of the IEEE/CVF conference on computer vision and pattern recognition}, 2022, pp. 13232--13242.

\bibitem{huang2017arbitrary}
Xun Huang and Serge Belongie,
\newblock ``Arbitrary style transfer in real-time with adaptive instance normalization,''
\newblock in {\em Proceedings of the IEEE international conference on computer vision}, 2017, pp. 1501--1510.

\bibitem{Choi2020Pose2MeshGC}
Hongsuk Choi, Gyeongsik Moon, et~al.,
\newblock ``Pose2mesh: Graph convolutional network for 3d human pose and mesh recovery from a 2d human pose,''
\newblock {\em ArXiv}, vol. abs/2008.09047, 2020.

\bibitem{uguz2024mocap}
Bedirhan Uguz, Ozhan Suat, et~al.,
\newblock ``Mocap-to-visual domain adaptation for efficient human mesh estimation from 2d keypoints,''
\newblock in {\em Proceedings of the IEEE/CVF Conference on Computer Vision and Pattern Recognition}, 2024, pp. 1622--1632.

\bibitem{wei2022capturing}
Wen-Li Wei, Jen-Chun Lin, et~al.,
\newblock ``Capturing humans in motion: Temporal-attentive 3d human pose and shape estimation from monocular video,''
\newblock in {\em Proceedings of the IEEE/CVF Conference on Computer Vision and Pattern Recognition}, 2022, pp. 13211--13220.

\bibitem{von2018recovering}
Timo Von~Marcard, Roberto Henschel, et~al.,
\newblock ``Recovering accurate 3d human pose in the wild using imus and a moving camera,''
\newblock in {\em Proceedings of the European conference on computer vision (ECCV)}, 2018, pp. 601--617.

\bibitem{ionescu2013human3}
Catalin Ionescu, Dragos Papava, et~al.,
\newblock ``Human3. 6m: Large scale datasets and predictive methods for 3d human sensing in natural environments,''
\newblock {\em IEEE transactions on pattern analysis and machine intelligence}, vol. 36, no. 7, pp. 1325--1339, 2013.

\bibitem{mehta2017monocular}
Dushyant Mehta, Helge Rhodin, et~al.,
\newblock ``Monocular 3d human pose estimation in the wild using improved cnn supervision,''
\newblock in {\em 2017 international conference on 3D vision (3DV)}. IEEE, 2017, pp. 506--516.

\bibitem{lin2014microsoft}
Tsung-Yi Lin, Michael Maire, et~al.,
\newblock ``Microsoft coco: Common objects in context,''
\newblock in {\em Computer Vision--ECCV 2014: 13th European Conference, Zurich, Switzerland, September 6-12, 2014, Proceedings, Part V 13}. Springer, 2014, pp. 740--755.

\bibitem{andriluka20142d}
Mykhaylo Andriluka, Leonid Pishchulin, et~al.,
\newblock ``2d human pose estimation: New benchmark and state of the art analysis,''
\newblock in {\em Proceedings of the IEEE Conference on computer Vision and Pattern Recognition}, 2014, pp. 3686--3693.

\end{thebibliography}

\vspace{12pt}

\end{document}